\documentclass{elsarticle}
\usepackage{xcolor}
\usepackage{url}
\usepackage{amsmath}
\usepackage{graphicx}

\makeatletter

\let\SF@@footnote\footnote
\def\footnote{\ifx\protect\@typeset@protect
    \expandafter\SF@@footnote
  \else
    \expandafter\SF@gobble@opt
  \fi
}
\expandafter\def\csname SF@gobble@opt \endcsname{\@ifnextchar[
  \SF@gobble@twobracket
  \@gobble
}
\edef\SF@gobble@opt{\noexpand\protect
  \expandafter\noexpand\csname SF@gobble@opt \endcsname}
\def\SF@gobble@twobracket[#1]#2{}
\providecommand{\tabularnewline}{\\}

\usepackage{amsmath,amsfonts,bm}

\def\eqref#1{equation~\ref{#1}}

\def\1{\bm{1}}

\DeclareMathAlphabet{\mathsfit}{\encodingdefault}{\sfdefault}{m}{sl}
\SetMathAlphabet{\mathsfit}{bold}{\encodingdefault}{\sfdefault}{bx}{n}

\usepackage[T1]{fontenc}
\usepackage[utf8]{inputenc}
\usepackage{xcolor}
\usepackage{array}
\usepackage{float}
\usepackage{textcomp}
\usepackage{url}
\usepackage{multirow}
\usepackage{amsmath}
\usepackage{amssymb}
\usepackage{graphicx}
\usepackage[draft]{todonotes}
\usepackage{microtype}
\usepackage{makecell}
\usepackage{tabu}
\usepackage[romanian, vietnamese, english]{babel}

\usepackage{tikz}
\usetikzlibrary{shapes.misc, positioning}
\usetikzlibrary{backgrounds,fit,matrix, calc}
\tikzset{
   box/.style = {minimum height=10pt, minimum width=10pt, draw, rounded corners,rectangle, fill=white!50},
}
\usetikzlibrary{fadings}

\tikzset{
   boxconv/.style = {minimum height=2cm, minimum width=2cm, draw, , fill opacity=0.9, rounded corners,rectangle, fill=white!50},
}

\tikzset{
   input/.style = {minimum height=3cm, minimum width=3cm, draw, , fill opacity=0.9, rectangle, fill=white!50},
}

\tikzset{
   boxpooled/.style = {minimum height=1.5cm, minimum width=1.5cm, draw, , fill opacity=0.9, rounded corners,rectangle, fill=white!50},
}

\usetikzlibrary{fadings}\tikzset{
    boxwta/.style={%
        draw=black, thick,
        rectangle,
        rounded corners,
        minimum height=3cm,
        minimum width=3cm
    }
}

\usetikzlibrary{fadings}\tikzset{
    box1/.style={%
        draw=black, thick,
        rectangle,
        minimum height=2cm,
        minimum width=2cm
    }
}

\usetikzlibrary{fadings}\tikzset{
    box2/.style={%
        draw=black, thick,
        rectangle,
        minimum height=1.cm,
        minimum width=1.cm
    }
}

\usetikzlibrary{fadings}\tikzset{
    box3/.style={%
        draw=black, thick,
        rectangle,
        minimum height=.8cm,
        minimum width=.8cm
    }
}

\usepackage{ifthen}

\usepackage{url}
\usepackage{caption}
\usepackage{nicefrac}
\usepackage{tabularx}

\@ifundefined{showcaptionsetup}{}{%
 \PassOptionsToPackage{caption=false}{subfig}}
\usepackage{subfig}

\begin{document}
\title{Amortized Context Vector Inference for Sequence-to-Sequence Networks}
\author{Kyriakos Tolias, Ioannis Kourouklides, and Sotirios Chatzis}
\address{Department of Electrical Eng., Computer Eng., and Informatics\\Cyprus University of Technology\\33, Saripolou Str.\\Limassol 3036, Cyprus}
\ead{ioannis@kourouklides.com, k.tolias@gmail.com, sotirios.chatzis@cut.ac.cy}
 
\begin{abstract}
Neural attention (NA) has become a key component of sequence-to-sequence
models that yield state-of-the-art performance in as hard tasks as
abstractive document summarization (ADS) and video captioning (VC). 
NA mechanisms perform inference of context
vectors; these constitute weighted sums of \emph{deterministic} input
sequence \emph{encodings}, adaptively sourced over long temporal horizons.
Inspired from recent work in the field of amortized variational inference
(AVI), in this work we consider treating the \emph{context vectors}
generated by soft-attention (SA) models as latent variables, with
approximate \emph{finite mixture model} posteriors inferred via AVI.
We posit that this formulation may yield stronger generalization capacity,
in line with the outcomes of existing applications of AVI to deep
networks. To illustrate our method, we implement it and experimentally
evaluate it considering challenging ADS and VC benchmarks. This
way, we exhibit its improved effectiveness over state-of-the-art alternatives. 
\end{abstract}

\begin{keyword}
Sequence-to-sequence, neural attention, variational inference, natural
language. 
\end{keyword}

\maketitle

\section{Introduction}

Sequence-to-sequence (\emph{seq2seq}) or encoder-decoder models \cite{Sutskever2014}
constitute a novel solution to inferring relations between sequences
of different lengths. They are broadly used for addressing tasks including
abstractive document summarization (ADS), descriptive caption generation (DCG)
\cite{Xu2015}, and question answering (QA) \cite{Sukhbaatar2015},
to name just a few. \emph{Seq2seq} models comprise two distinct RNN
models: an \emph{encoder }RNN, and a \emph{decoder }RNN. Their main
principle of operation is based on the idea of learning to infer an
intermediate \emph{context vector representation, $\boldsymbol{c}$,
}which is ``shared'' among the two RNN modules of the model, i.e.,
the encoder and the decoder. Specifically, the encoder converts the
source sequence to a context vector (e.g., the final state of the
encoder RNN), while the decoder is presented with the inferred context
vector to produce the target sequence.

Despite these merits, though, baseline \emph{seq2seq} models cannot
learn temporal dynamics over long horizons. This is due to the fact
that a single context vector $\boldsymbol{c}$ is capable of encoding
rather limited temporal information. This major limitation has been
addressed via the development of neural attention (NA) mechanisms
\cite{Bahdanau2014}. NA has been a major breakthrough in Deep Learning
for Natural Language Processing, as it enables the decoder modules
of \emph{seq2seq} models to adaptively focus on temporally-varying
subsets of the source sequence. This capacity, in turn, enables flexibly
capturing long temporal dynamics in a computationally efficient manner.

Among the large collection of recently devised NA variants, the vast
majority build upon the concept of \emph{Soft Attention} (SA) \cite{Xu2015}.
Under this rationale, at each sequence generation (decoding) step,
NA-obtained context vectors essentially constitute deterministic representations
of the dynamics between the source sequence and the decodings obtained
thus far. However, recent work in the field of amortized variational
inference (AVI) \cite{JimenezRezende2015,Kingma2013} has shown that
it is often useful to treat representations generated by deep networks
as \emph{latent random }variables\emph{. }Indeed, it is now well-understood
that, under such an inferential setup, the trained deep learning models
become more effective in inferring representations that offer stronger
generalizaton capacity, instead of getting trapped to representations
of poor generalizaton quality. Then, model training reduces to inferring
posterior distributions over the introduced latent variables. This
can be performed by resorting to variational inference \cite{Attias2000},
where the sought variational posteriors are parameterized via appropriate
deep networks.

Motivated from these research advances, in this paper we consider
a novel formulation of SA. Specifically, we propose an NA mechanism
formulation where the generated context vectors are considered random
latent variables with \emph{finite mixture model posteriors}, over
which AVI is performed. We dub our approach amortized context vector
inference (ACVI). To exhibit the efficacy of ACVI, we implement it
into: (i) Pointer-Generator Networks \cite{Manning2017}, which constitute
a state-of-the-art approach for addressing ADS tasks; and (ii) baseline
\emph{seq2seq} models with additive SA, applied to the task of VC.

The remainder of this paper is organized as follows: In Section II,
we briefly present the \emph{seq2seq }model variants in the context
of which we implement our method and exhibit its efficacy. In Section
III, we introduce the proposed approach, and elaborate on its training
and inference algorithms. In Section IV, we perform an extensive experimental
evaluation of our approach using benchmark ADS and VC datasets.
Finally, in the concluding Section, we summarize the contribution
of this work.

\section{Methodological Background}

\subsection{Abstractive Document Summarization}

ADS consists in not only copying from an original document, but also
learning to generate new sentences or novel words during the summarization
process. The introduction of \emph{seq2seq} models has rendered ADS
both feasible and effective \cite{Rush2015a,Zeng2016}. Dealing with
out-of-vocabulary (OOV) words was one of the main difficulties that
early ADS models were confronted with. Word and/or phrase repetition
was a second issue. The \emph{pointer-generator} model presented in
\cite{Manning2017} constitutes one of the most comprehensive efforts
towards ameliorating these issues.

In a nutshell, this model comprises one bidirectional LSTM \cite{Hochreiter1997}
(BiLSTM) encoder, and a unidirectional LSTM decoder, which incorporates
an SA mechanism \cite{Bahdanau2014}. The word embedding of each token,
$\boldsymbol{x}_{i},\;i\in\{1,\dots,N\}$, in the source sequence
(document) is presented to the encoder BiLSTM; this obtains a representation
(encoding) $\boldsymbol{h}_{i}=[\overrightarrow{\boldsymbol{h}}_{i};\overleftarrow{\boldsymbol{h}}_{i}]$,
where $\overrightarrow{\boldsymbol{h}}_{i}$ is the corresponding
forward LSTM state, and $\overleftarrow{\boldsymbol{h}}_{i}$ is the
corresponding backward LSTM state. Then, at each generation step,
$t$, the decoder LSTM gets as input the (word embedding of the) previous
token in the target sequence. During training, this is the previous
word in the available reference summary; during inference, this is
the previous generated word. On this basis, the decoder updates its
internal state, $\boldsymbol{s}_{t}$, which is then presented to
the postulated SA network. Specifically, the attention distribution,
$\boldsymbol{a}_{t}$, is given by: 
\begin{equation}
e_{t}^{i}=\boldsymbol{v}^{T}\tanh(\boldsymbol{W}_{h}\boldsymbol{h}_{i}+\boldsymbol{W}_{s}\boldsymbol{s}_{t}+\boldsymbol{b}_{attn})\label{eq:attention}
\end{equation}
\begin{equation}
\boldsymbol{a}_{t}=\mathrm{softmax}(\boldsymbol{e}_{t}),\;\boldsymbol{e}_{t}=[e_{t}^{i}]_{i}\label{eq:alignments}
\end{equation}
where the $\boldsymbol{W}_{\cdot}$ are trainable weight matrices,
$\boldsymbol{b}_{attn}$ is a trainable bias vector, and $\boldsymbol{v}$
is a trainable parameter vector of the same size as $\boldsymbol{b}_{attn}$.
Then, the model updates the maintained context vector, $\boldsymbol{c}_{t}$,
by taking an weighted average of all the source token encodings; in
that average, the used weights are the inferred attention probabilities.
We obtain: 
\begin{equation}
\boldsymbol{c}_{t}=\sum_{i}a_{t}^{i}\boldsymbol{h}_{i}\label{eq:context}
\end{equation}
Eventually, the \emph{predictive distribution} over the next \emph{generated}
word yields: 
\begin{equation}
P_{t}^{vocab}=\mathrm{softmax(}\boldsymbol{V}'\mathrm{tanh}(\boldsymbol{V}[\boldsymbol{s}_{t};\boldsymbol{c}_{t}]+\boldsymbol{b})+\boldsymbol{b}^{'})\label{eq:pred_distr}
\end{equation}
where $\boldsymbol{V}$ and $\boldsymbol{V}'$ are trainable weight
matrices, while $\boldsymbol{b}$ and $\boldsymbol{b}'$ are trainable
bias vectors.

In parallel, the network also computes an additional probability,
$p_{t}^{gen}$, which expresses whether the next output should be
\emph{generated} by sampling from the predictive distribution, $P_{t}^{vocab}$,
or the model should simply \emph{copy} one of the already available
s\emph{ource sequence tokens}. This mechanism allows for the model
to cope with OOV words; it is defined via a simple sigmoid layer of
the form: 
\begin{equation}
p_{t}^{gen}=\sigma(\boldsymbol{w}_{c}^{T}\boldsymbol{c}_{t}+\boldsymbol{w}_{s}^{T}\boldsymbol{s}_{t}+\boldsymbol{w}_{x}^{T}\boldsymbol{x}_{t}+\boldsymbol{b}_{ptr})
\end{equation}
where $\boldsymbol{x}_{t}$ is the decoder input, while the $\boldsymbol{w}_{\cdot}$
and $\boldsymbol{b}_{ptr}$ are trainable parameter vectors. The probability
of copying the $i$th source sequence token is considered equal to
the corresponding attention probability, $a_{t}^{i}$. Eventually,
the obtained probability that the next output word will be $\beta$
(found either in the vocabulary or among the source sequence tokens)
yields: 
\begin{equation}
P_{t}(\beta)=p_{t}^{gen}P_{t}^{vocab}(\beta)+(1-p_{t}^{gen})\sum_{i:\beta_{i}=\beta}a_{t}^{i}
\end{equation}
Finally, a \emph{coverage mechanism}\textbf{ }may also be employed
\cite{Tu2016}, as a means of penalizing words that have already received
attention in the past, to prevent repetition. Specifically, the coverage
vector, $\boldsymbol{k}_{t}$, is defined as: 
\begin{equation}
\boldsymbol{k}_{t}=[k_{t}^{i}]_{i=1}^{N}=\sum_{\tau=0}^{t-1}\boldsymbol{a}_{\tau}
\end{equation}
Using the so-obtained coverage vector, expression (\ref{eq:attention})
is modified as follows: 
\begin{equation}
e_{t}^{i}=\boldsymbol{v}^{T}\tanh(\boldsymbol{W}_{h}\boldsymbol{h}_{i}+\boldsymbol{W}_{s}\boldsymbol{s}_{t}+\boldsymbol{w}_{k}k_{t}^{i}+\boldsymbol{b}_{attn})\label{eq:attention-1}
\end{equation}
where $\boldsymbol{w}_{k}$ is a trainable parameter vector of size
similar to $\boldsymbol{v}$. Model training is performed via minimization
of the categorical cross-entropy, augmented with a coverage term of
the form: 
\begin{equation}
\lambda\sum_{i}\sum_{t}\mathrm{min}(a_{t}^{i},c_{t}^{i})
\end{equation}
Here, $\lambda$ controls the influence of the coverage term; in the
remainder of this work, we set $\lambda=1$.

\subsection{Video Captioning}

\label{VideoCaptioning}

\emph{Seq2seq} models with attention have been successfully applied
to several datasets of multimodal nature. Video captioning constitutes
a popular such application. In this work, we consider a simple\emph{
seq2seq }model with additive SA that comprises a BiLSTM encoder, an
LSTM decoder, and an output distribution of the form (4). The used
encoder is presented with visual features obtained from a \emph{pretrained}
convolutional neural network (CNN). Using a pretrained CNN as our
employed visual feature extractor ensures that all the evaluated attention
models are presented with identical feature descriptors of the available
raw data. Hence, it facilitates fairness in the comparative evaluation
of our proposed attention mechanism. We elaborate on the specific
model configuration in Section IV.B.

\section{Proposed Approach}

\label{approach}

We begin by introducing the core assumption that the computed context
vectors, $\boldsymbol{c}_{t}$, constitute latent random variables.
Further, we assume that, at each decoding step, $t$, the corresponding
context vector, $\boldsymbol{c}_{t}$, is drawn from a distribution
associated with one of the available source sequence encodings, $\{\boldsymbol{h}_{i}\}_{i=1}^{N}$.
The selection of the source sequence encoding to associate with is
determined from the output sequence via the decoder state, $\boldsymbol{s}_{t}$,
as we explain next.

Let us introduce the set of binary latent indicator variables, $\{z_{t}^{i}\}_{i=1}^{N}$,
$z_{t}^{i}\in\{0,1\}$, with $z_{t}^{i}=1$ denoting that the context
vector $\boldsymbol{c}_{t}$ is drawn from the $i$th density, that
is the density associated with the $i$th source encoding, $\boldsymbol{h}_{i}$,
and $z_{t}^{i}=0$ otherwise. Then, we postulate the following hierarchical
model: 
\begin{equation}
\boldsymbol{c}_{t}|z_{t}^{i}=1;\mathcal{D}\sim p(\boldsymbol{\theta}(\boldsymbol{h}_{i}))\label{eq:cond_posterior}
\end{equation}
\begin{equation}
z_{t}^{i}=1|\mathcal{D}\sim\pi_{t}^{i}(a_{t}^{i})\label{eq:ind_posterior}
\end{equation}
where $\mathcal{D}$ comprises the set of \emph{both the source and
target} training sequences, $\boldsymbol{\theta}$ denotes the parameters
set of the context vector conditional density, and $\pi_{t}^{i}$
denotes the probability of drawing from the $i$th conditional at
time $t$. Notably, we assume that the component assignment probabilities,
$\pi_{t}^{i}$, are functions of the attention probabilities, $a_{t}^{i}$.
Thus, the selection of the mixture component density that we draw
the context vector from at decoding time $t$ is directly determined
from the value of the current decoder state, $\boldsymbol{s}_{t}$,
via the corresponding attention probabilities. A higher affinity of
the current decoder state $\boldsymbol{s}_{t}$ with the $i$th encoding,
$\boldsymbol{h}_{i}$, at time $t$, results in higher probability
that the context vector be drawn from the corresponding conditional
density.

Having defined the hierarchical model (\ref{eq:cond_posterior})-(\ref{eq:ind_posterior}),
it is important that we examine the resulting expression of the posterior
density $p(\boldsymbol{c}_{t};\mathcal{D})$. By marginalizing over
(\ref{eq:cond_posterior}) and (\ref{eq:ind_posterior}), we obtain:
\begin{equation}
p(\boldsymbol{c}_{t};\mathcal{D})=\sum_{i=1}^{N}\pi_{t}^{i}(a_{t}^{i})\,p(\boldsymbol{\theta}(\boldsymbol{h}_{i}))\label{eq:generic}
\end{equation}
In other words, we obtain a\emph{ finite mixture model posterior over
the context vectors,} with mixture conditional densities associated
with the available source sequence encodings, and mixture weights
that are functions of the corresponding attention vectors, and are
therefore determined by the target sequences.

In addition, it is interesting to compare this expression to the definition
of context vectors under the conventional SA scheme. From (\ref{eq:context}),
we observe that conventional SA is merely a special case of our proposed
model, obtained by introducing two assumptions: (i) that the postulated
mixture component assignment probabilities are identity functions
of the associated attention probabilities, i.e. 
\begin{equation}
\begin{aligned}p(\boldsymbol{z}_{t};\mathcal{D})=&\mathrm{Cat}(\boldsymbol{z}_{t}\big|\boldsymbol{\pi}_{t}),\; \boldsymbol{z}_{t}=[z_{t}^{i}]_{i=1}^{N},\boldsymbol{\pi}_{t}=[\pi_{t}^{i}(a_{t}^{i})]_{i=1}^{N}\\
 & \mathrm{s.t.\quad}\pi_{t}^{i}(a_{t}^{i})\triangleq p(z_{t}^{i}=1;\mathcal{D})=a_{t}^{i}=\mathrm{softmax}(\boldsymbol{e}_{t});\label{eq:mixing}
\end{aligned}
\end{equation}
and (ii) that the conditional densities of the context vectors have
all their mass concentrated on $\boldsymbol{h}_{i}$, that is they
collapse onto the single point, $\boldsymbol{h}_{i}$: 
\begin{equation}
p(\boldsymbol{c}_{t}|z_{t}^{i}=1;\mathcal{D})=\delta(\boldsymbol{h}_{i})\label{eq:delta}
\end{equation}
Indeed, by combining (\ref{eq:generic}) - (\ref{eq:delta}), we yield:
\begin{equation}
p(\boldsymbol{c}_{t};\mathcal{D})=\sum_{i=1}^{N}a_{t}^{i}\delta(\boldsymbol{h}_{i})
\end{equation}
whence we obtain (\ref{eq:context}) with probability 1.

Thus, our approach replaces the simplistic conditional density expression
(\ref{eq:delta}) with a more appropriate family $p(\boldsymbol{\theta}(\boldsymbol{h}_{i}))$,
as in (\ref{eq:generic}). Based on the literature of AVI, e.g. \cite{JimenezRezende2015,Kingma2013,ladder},
we posit that such a \emph{stochastic} latent variable consideration
may result in significant advantages for the postulated \emph{seq2seq}
model. Specifically, our trained model becomes more agile in searching
for effective context representations, as opposed to getting trapped
to poor local solutions.

In the following, we examine conditional densities of Gaussian form.
Adopting the inferential rationale of AVI, we consider that these
conditional Gaussians are parameterized via the postulated BiLSTM
encoder. Specifically, we assume: 
\begin{equation}
p(\boldsymbol{c}_{t}|z_{t}^{i}=1;\mathcal{D})=\mathcal{N}\big(\boldsymbol{c}_{t}|\boldsymbol{h}_{i},\mathrm{diag}(\boldsymbol{\sigma}^{2}(\boldsymbol{h}_{i}))\big)\label{eq:Gaussian}
\end{equation}
where 
\begin{equation}
\mathrm{log}\,\boldsymbol{\sigma}^{2}(\boldsymbol{h})=\mathrm{ReLU}(\boldsymbol{h})\label{eq:mean}
\end{equation}
$\mathrm{ReLU}(\cdot)$ is a trainable ReLU layer of size $\mathrm{dim}(\boldsymbol{h})$,
and the encodings, $\boldsymbol{h}_{i}$, are obtained from a BiLSTM
encoder, similar to conventional models. Hence: 
\begin{equation}
p(\boldsymbol{c}_{t};\mathcal{D})=\sum_{i=1}^{N}a_{t}^{i}\mathcal{N}\big(\boldsymbol{c}_{t}|\boldsymbol{h}_{i},\mathrm{diag}(\boldsymbol{\sigma}^{2}(\boldsymbol{h}_{i}))\big)\label{eq:cond_Gaussian}
\end{equation}

Thus, we have arrived at an approximate (variational) posterior expression
for the context vectors, $\boldsymbol{c}_{t}$. In our variational
treatment, both the component-conditional means, $\boldsymbol{h}_{i}$,
and their variances, $\boldsymbol{\sigma}^{2}(\boldsymbol{h}_{i})$,
are obtained from (amortizing) neural networks presented with the
source sequences. On the other hand, though, the assignment probabilities,
$\pi_{t}^{i}$, in the variational posterior are taken as the attention
probabilities, $a_{t}^{i}$. Thus, they are determined by the target
sequences, which are generated from the decoder of the model. Hence,
our treatment represents a valid approximate posterior formulation,
overall conditioned on both the source and target sequences.

This concludes the formulation of ACVI.

\textbf{Relation to Recent Work.} From the above exhibition, it becomes
apparent that our approach generalizes the concept of neural attention
by introducing stochasticity in the computation of context vectors.
As we have already discussed, the ultimate goal of this construction
is to allow for inferring representations of better generalization
capacity, by leveraging Bayesian inference arguments.

We emphasize that this is in stark contrast to recent efforts toward
generalizing neural attention by deriving more complex attention distributions.
For instance, \cite{structuredAttention} have recently introduced
\emph{structured attention}. In that work, the model infers complex
posterior probabilities over the assignment latent variables, as opposed
to using a simplistic gating function. Specifically, instead of considering
independent assignments, they postulate the first-order Markov dynamics
assumption: 
\begin{equation}
p(\{\boldsymbol{z}_{t}\}_{t=1}^{T};\mathcal{D})=p(\boldsymbol{z}_{1};\mathcal{D})\prod_{t=1}^{T-1}p(\boldsymbol{z}_{t+1}|\boldsymbol{z}_{t};\mathcal{D})\label{eq:Structured}
\end{equation}
Thus, \cite{structuredAttention} compute posterior distributions
over the attention assignments, while ACVI provides a method for obtaining
improved representations through the inferred context vectors. Note
also that Eq. (\ref{eq:Structured}) gives rise to the need of executing
much more computationally complex algorithms to perform attention
distribution inference, e.g. the forward-backward algorithm \cite{rabiner}.
In contrast, our method imposes computational costs comparable to
conventional SA.

Similar is the innovation in the \emph{variational attention} method,
recently presented \cite{nips2018}. In essence, its key conceptual
difference from \emph{structured attention} is the consideration of
full independence between the attention assignments $\{\boldsymbol{z}_{t}\}_{t=1}^{T}$.
Among the several alternatives considered in \cite{nips2018} to obtain
stochastic gradient estimators of low variance, it was found that
an approach using REINFORCE \cite{reinforce} along with a specialized
baseline was effective.

Another noteworthy recent work, closer related to ACVI, is the variational
encoder-decoder (VED) method presented in \cite{variational_att}.
Among the several alternative formulations considered in that paper,
the one that clearly outperformed the baselines in terms of the obtained
accuracy (BLEU scores) combined \emph{seq2seq} models with SA with
an extra variational autoencoder (VAE) module. This way, apart from
the context vector, which is computed under the standard SA scheme,
an additional latent vector $\boldsymbol{\xi}$ is essentially inferred.
The imposed prior over it is a standard $\mathcal{N}(\boldsymbol{0},\boldsymbol{I})$,
while the inferred posterior is a diagonal Gaussian parameterised
by a BiLSTM network presented with the input sequence; the final BiLSTM
state vector is presented to dense layers that output the posterior
means and variances of the latent vectors $\boldsymbol{\xi}$. Both
the context vector, $\boldsymbol{c}$, as well as the latent vectors,
$\boldsymbol{\xi}$, are fed to the final softmax layer of the model
that yields the generated output symbols.

We shall provide comparisons to all these related approaches in the
experimental section of our paper.

\textbf{Training Algorithm.} To perform training of a \emph{seq2seq
}model equipped with the ACVI mechanism, we resort to maximization
of the resulting evidence lower-bound (ELBO) expression. To this end,
we need first to introduce some prior assumption over the context
latent variables, $\boldsymbol{c}_{t}$. To serve the purpose of simplicity,
and also offer a valid way to effect model regularization, we consider:
\begin{equation}
p(\boldsymbol{c}_{t})=\mathcal{N}\big(\boldsymbol{c}_{t}|\boldsymbol{0},\boldsymbol{I})
\end{equation}

On the grounds of these assumptions, it is easy to show that the resulting
ELBO expression becomes: 
\begin{equation}
\mathcal{L}=\sum_{t}\left\{ \mathbb{E}_{p(\boldsymbol{c}_{t};\mathcal{D})}[-J_{t}]\,-\mathrm{KL}[p(\boldsymbol{c}_{t};\mathcal{D})||p(\boldsymbol{c}_{t})]\right\} \label{eq: ELBO}
\end{equation}
In this expression, $\mathbb{E}_{p(\boldsymbol{c}_{t};\mathcal{D})}[-J_{t}]$
is the posterior expectation of the model log-likelihood, which is
an integral part of the ELBO definition. In the following, we approximate
all the entailed ELBO terms by drawing MC samples from the context
vector posterior. In this work, we are dealing with a one-out-of-many
predictive selection; hence, the model likelihood is a simple Categorical.
As such, $\mathbb{E}_{p(\boldsymbol{c}_{t};\mathcal{D})}[-J_{t}]$
essentially reduces to the negative categorical cross-entropy of the
model, averaged over multiple MC samples of the context vectors, drawn
from (19). Besides, to ensure that the resulting MC estimators will
be of low variance, we adopt the reparameterization trick. To this
end, we rely on the posterior expressions (\ref{eq:Gaussian}) and
(\ref{eq:mixing}); we express the drawn MC samples as follows: 
\begin{equation}
\boldsymbol{c}_{t}^{(k)}=\sum_{i=1}^{N}z_{ti}^{(k)}\boldsymbol{c}_{ti}^{(k)}\label{sampling}
\end{equation}

In this expression, the $\boldsymbol{c}_{ti}^{(k)}$ are samples from
the conditional Gaussians (\ref{eq:Gaussian}), which employ the standard
reparameterization trick rationale, as applied to Gaussian variables:
\begin{equation}
\boldsymbol{c}_{ti}^{(k)}=\boldsymbol{h}_{i}+\boldsymbol{\sigma}(\boldsymbol{h}_{i})\circ\boldsymbol{\epsilon}_{ti}^{(k)},\;\boldsymbol{\epsilon}\sim\mathcal{N}(\boldsymbol{0},\boldsymbol{I})
\end{equation}
On the other hand, the $z_{ti}^{(k)}$ are samples from the Categorical
distribution (\ref{eq:mixing}). To allow for performing backpropagation
through these samples, while ensuring that the obtained gradients
will be of low variance, we may draw $z_{ti}^{(k)}$ by making use
of the Gumbel-Softmax relaxation \cite{gumbel}. We have empirically
found it suffices that we employ the Gumbel-Softmax trick for the
last 10\% of the model training iterations\footnote{We use a Gumbel-Softmax temperature hyperparameter value of 0.5.};
previously, we merely adopt the following heuristic, without any statistically
significant performance deviation: We use a simple weighted average
of the samples $\boldsymbol{c}_{ti}^{(k)}$, with the weights being
the attention probabilities, $a_{t}^{i}$: 
\begin{equation}
\boldsymbol{c}_{t}^{(k)}\leftarrow\sum_{i=1}^{N}a_{t}^{i}\boldsymbol{c}_{ti}^{(k)}\label{approxsampling}
\end{equation}
This way, we alleviate the computational costs of employing the Gumbel-Softmax
relaxation, which dominates the costs of sampling from the mixture
posterior (\ref{eq:cond_Gaussian}).

Having obtained a reparameterization of the model ELBO that guarantees
low variance estimators, we proceed to its maximization by resorting
to a modern, off-the-shelf, stochastic gradient optimizer. Specifically,
we adopt Adam with its default settings \cite{adam}.

\textbf{Inference Algorithm.} To perform target decoding by means
of a \emph{seq2seq }model that employs the ACVI mechanism, we resort
to \emph{Beam search }\cite{Russel}. In our experiments, \emph{Beam}
\emph{width} is set to five.


\begin{table*}
\caption{Abstractive Document Summarization: ROUGE scores on the test set.}

\definecolor{mint}{RGB}{107,142,35}

\centering{}\small%
\begin{tabular}{|c|c|c|c|c|c|}
\hline 
\multirow{2}{*}{Method} & \multicolumn{3}{c|}{ROUGE} & \multicolumn{2}{c|}{METEOR}\tabularnewline
\cline{2-6} 
 & 1 & 2 & L & Exact Match & + stem/syn/para\tabularnewline
\hline 
\emph{seq2seq }with SA & 31.33 & 11.81 & 28.83 & 12.03 & 13.20\tabularnewline 
\cline{2-6} 
pointer-generator + coverage: \textbf{SA} & {39.53} & {17.28} & {36.38} & {17.32} & {18.72}
\tabularnewline
\hline
\makecell{pointer-generator + coverage: \\  \textbf{structured attention}} & {40.12} & {17.61} & {36.74} & {17.38} & {18.93}
\tabularnewline
\hline 
\makecell{pointer-generator + coverage: \\  \textbf{variational attention}}  & 40.04 & 17.37 & 36.45 & 17.14 & 18.66 \tabularnewline
\hline 
pointer-generator + coverage: \textbf{VED} & 41.28 & 18.05 & 38.12 & 17.63 & 18.87 \tabularnewline
\hline
pointer-generator + coverage: \textbf{ACVI} & \textbf{{42.71}} & \textbf{{19.24}} & \textbf{{39.05}} & \textbf{{18.47}} & \textbf{{20.09}}\tabularnewline
\hline
\end{tabular}
\end{table*}

\begin{table*}
\centering{}\caption{Abstractive Document Summarization: Novel words generation rate and
OOV words adoption rate obtained by using pointer-generator networks.}
\small%
\begin{tabular}{|c|c|c|c|c|c|}
\cline{2-6} 
\multicolumn{1}{c|}{} & SA & Structured Attention & Variational Attention & VED & ACVI\tabularnewline
\hline 
Rate of Novel Words & 0.05 & 0.05 & 0.05 & 0.12 & 0.38\tabularnewline
\hline 
Rate of OOV Words Adoption & 1.16 & 1.18 & 1.18 & 1.21 & 1.25\tabularnewline
\hline 
\end{tabular}
\end{table*}

\section{Experimental Evaluation\protect\footnote{We have developed our source codes in Python, using the TensorFlow
library \cite{tensorflow2015-whitepaper}. }}

\subsection{Abstractive Document Summarization}

\label{ADS}

Our experiments are based on the \textit{non-anonymized} \textit{CNN/Daily
Mail} dataset, similar to the experiments of \cite{Manning2017}.
To obtain some comparative results, we use pointer-generator networks
as our evaluation platform \cite{Manning2017}; therein, we employ
our ACVI mechanism, the standard SA mechanism used in \cite{Manning2017},
VED \cite{variational_att}, variational attention \cite{nips2018},
as well as structured attention using the first-order Markov assumption
(\ref{eq:Structured}) \cite{structuredAttention}. The observations
presented to the encoder modules constitute 128-dimensional word embeddings
of the original 50K-dimensional one-hot-vectors of the source tokens.
Similarly, the observations presented to the decoder modules are 128-dimensional
word embeddings pertaining to the summary tokens (reference tokens
during training; generated tokens during inference). Both these embeddings
are trained, as part of the overall training procedure of the evaluated
models. To allow for faster training convergence, we split training
into five phases, as suggested in \cite{Manning2017}. Following the
suggestions in \cite{Manning2017}, we evaluate all approaches with
LSTMs that comprise 256-dimensional states and do not employ Dropout.
We have tested VED with various selections of the dimensionality of
the autoencoder latent vectors, $\boldsymbol{\xi}$; we report results
with 128-dimensional latent vectors, which yielded the best performance
in our experiments\footnote{This selection is also commensurate with the $\boldsymbol{\xi}$ dimensionality
reported in \cite{variational_att}.}.

Finally, for completeness sake, we also evaluate the Transformer network
\cite{transformer}, which is a popular alternative to \emph{seq2seq}
models with SA, based on the notion of self-attention. Following \cite{web},
Transformer is evaluated with 256-dimensional word embeddings, 4 encoding
and decoding layers of 256 units each, 4 heads, and a Dropout rate
of 0.2.

We use ROUGE\footnote{\url{pypi.python.org/pypi/pyrouge/}.} \cite{Lin2004}
and METEOR\footnote{\url{www.cs.cmu.edu/~alavie/METEOR}.} \cite{Denkowski2014}
as our performance metrics. METEOR is evaluated both in exact match
mode (rewarding only exact matches between words) and full mode (additionally
rewarding matching stems, synonyms and paraphrases). In all our experiments,
we restrict the used vocabulary to the 50K most common words in the
considered dataset, similar to \cite{Manning2017}. Note that this
is significantly smaller than typical in the literature \cite{Nallapati2016}.
Our quantitative evaluation is provided in Table 1. Some indicative
examples of generated summaries can be found in Appendix A.

As we observe, utilization of ACVI outperforms all the alternatives
by a large margin. It is also interesting that the Transformer network
yields the lowest performance among the considered alternatives; the
obtained results are actually very poor. This is commensurate with
the results reported by other researchers, e.g. \cite{web}.

Finally, it is interesting to examine whether ACVI increases the propensity
of a trained model towards generating \emph{novel words, }that is
words that \emph{are not found }in the source document, as well as
the capacity to adopt OOV words. The related results are provided
in Table 2. We observe that ACVI increases the number of generated
novel words by 3 times compared to the best performing alternative,
that is VED \cite{variational_att}. In a similar vein, ACVI appears
to help the model better cope with OOV words.

\begin{table*}[htp]
\caption{Video Captioning: Performance of the considered alternatives.}

\centering{}{\small{}{}}%
\begin{tabular}{|c|c|c|c|c|}
\hline 
{\small{}{}Method }  & {\small{}{}ROUGE: Valid. Set }  & {\small{}{}ROUGE: Test Set }  & {\small{}{}CIDEr: Valid. Set }  & {\small{}{}CIDEr: Test Set}\tabularnewline
\hline 
{\small{}{}SA }  & {\small{}{}0.5628 }  & {\small{}{}0.5701 }  & {\small{}{}0.4575 }  & {\small{}{}0.421}\tabularnewline
\hline 
{\small{}{}Structured Attention }  & {\small{}{}0.5804 }  & {\small{}{}0.5712 }  & {\small{}{}0.5071 }  & {\small{}{}0.4283}\tabularnewline
\hline 
{\small{}{}Variational Attention }  & {\small{}{}0.5809 }  & {\small{}{}0.5716 }  & {\small{}{}0.5103 }  & {\small{}{}0.4289 }\tabularnewline
\hline 
{\small{}{}VED }  & {\small{}{}0.5839 }  & {\small{}{}0.5749 }  & {\small{}{}0.5421 }  & {\small{}{}0.4298 }\tabularnewline
\hline 
{\small{}{}ACVI }  & \textbf{\small{}{}0.5968}{\small{}{} }  & \textbf{\small{}{}0.5766}{\small{}{} }  & \textbf{\small{}{}0.6039}{\small{}{} }  & \textbf{\small{}{}0.4375}\tabularnewline
\hline 
\end{tabular}

\end{table*}

\subsection{Video Captioning}

Our evaluation of the proposed approach in the context of
a VC application is based on the Youtube2Text video corpus \cite{vc}.
We split the available dataset into a training set comprising the
first 1,200 video clips, a validation set composed of 100 clips, and
a test set comprising the last 600 clips in the dataset. To reduce
the entailed memory requirements, we process only the first 240 frames
of each video. To obtain some initial video frame descriptors, we
employ a pretrained GoogLeNet CNN \cite{googlenet} (implementation
provided in Caffe \cite{caffe}). Specifically, we use the features
extracted at the \emph{pool5/7x7\_s1 layer} of this
pretrained model. We select 24 equally-spaced frames out of the first
240 from each video, and feed them into the prescribed CNN to obtain
a 1024 dimensional frame-wise feature vector. These are the visual
inputs presented to the trained models. All employed LSTMs entail
1000-dimensional states. These are mapped to 100-dimensional features
via the matrices $\boldsymbol{W}_{h}$ and $\boldsymbol{W}_{s}$ in
Eq. (\ref{eq:attention}). The autoencoder latent variables, $\boldsymbol{\xi}$,
of VED are also selected to be 100-dimensional vectors. The decoders
are presented with 256-dimensional word embeddings, obtained in a
fashion similar to our ADS experiments. In all cases, we use Dropout
with a rate of 0.5.

We yield some comparative results by evaluating \emph{seq2seq} models 
configured as described in Section II.B; we use ACVI,
structured attention in the form (\ref{eq:Structured}), VED, variational
attention, or the conventional SA mechanism. Our quantitative evaluation
is performed on the grounds of the ROUGE-L and CIDEr \cite{cider}
scores, on both the validation set and the test set. The obtained
results are depicted in Table 3; they show that our method outperforms
the alternatives by an important margin. It is also characteristic
that Structured Attention yields essentially identical results with
Variational Attention. Thus, the first-order Markovian assumption
does not offer practical benefits when generating short sequences
like the ones involved in VC. Finally, we provide some indicative
examples of the generated results in Appendix B (Figs. 1-8).

\begin{table*}[htp]
\begin{center}
\caption{Abstractive Document Summarization: Domain Adaptation Performance on DUC2004.}
\begin{tabular}{|c|c|c|c|}
\hline 
{\small{}{}Model  } & {\small{}{}ROUGE-1  } & {\small{}{}ROUGE-2  } & {\small{}{}ROUGE-L}\tabularnewline
\hline 
\hline 
{\small{}{}SA  } & {\small{}{}27.02  } & {\small{}{}7.44  } & {\small{}{}22.69}\tabularnewline
\hline 
{\small{}{}Variational Attention  } & {\small{}{}27.65  } & {\small{}{}7.58  } & {\small{}{}23.50}\tabularnewline
\hline 
{\small{}{}VED  } & {\small{}{}30.68  } & {\small{}{}9.97  } & {\small{}{}27.02}\tabularnewline
\hline 
{\small{}{}ACVI  } & {\small{}{}32.09  } & {\small{}{}10.88  } & {\small{}{}28.14}\tabularnewline
\hline
\end{tabular}
\end{center}
\label{default}
\end{table*}%

\begin{table*}
\small%
\caption{Abstractive Document Summarization: Training phases.}
\centering{}%
\begin{tabular}{|c|c|c|c|}
\hline 
{\small{}{}Phase  } & {\small{}{}Iterations  } & {\small{}{}Max encod. steps  } & {\small{}{}Max decod. steps}\tabularnewline
\hline 
\hline 
{\small{}{}1  } & {\small{}{}0 - 71k  } & {\small{}{}10  } & {\small{}{}10}\tabularnewline
\hline 
{\small{}{}2  } & {\small{}{}71k - 116k  } & {\small{}{}50  } & {\small{}{}50}\tabularnewline
\hline 
{\small{}{}3  } & {\small{}{}116k - 184k  } & {\small{}{}100  } & {\small{}{}50}\tabularnewline
\hline 
{\small{}{}4  } & {\small{}{}184k - 223k  } & {\small{}{}200  } & {\small{}{}50}\tabularnewline
\hline 
{\small{}{}5  } & {\small{}{}223k - 250k  } & {\small{}{}400  } & {\small{}{}100}\tabularnewline
\hline 
\end{tabular}{\small\par}
\end{table*}

\subsection{Further Investigation: Domain Adaptation}

Finally, we wish to examine the capability of ACVI to generalize
across domains. We have already elaborated on our expectation that
modeling the context vectors as latent random variables should yield
improved generalization performance. We attribute to this fact the
improved accuracy ACVI obtained in our experimental evaluations. However,
if this is the case, one would probably expect the method to also
generalize better across different domains.

To investigate this aspect, we use the trained ADS models
described in Section IV.A to generate summaries for the documents of
the DUC2004 dataset . This is an English dataset comprising 500 documents.
Each document contains 4 model summaries written by experts. In Table 4, we show how our method performs in this setting, and
how it compares to the alternative variational methods considered
in Section IV.A. We observe that ACVI yields a clear improvement over
the alternatives, while all variational methods perform significantly
better than baseline SA. These findings seem to support our theoretical
intuitions.

\section{Conclusions}

In this work, we cast the problem of context vector computation
for \emph{seq2seq}-type models employing SA into
amortized variational inference. We made this possible by considering
that the sought context vectors are latent variables following a Gaussian
mixture posterior; therein, the mixture component densities depend
on the source sequence encodings, while the mixture weights depend
on the target sequence attention probabilities. We exhibited the merits
of our approach on \emph{seq2seq} architectures
addressing ADS and VC tasks; we used benchmark datasets in all
cases.

We underline that our approach induces only negligible computational
overheads compared to conventional SA. Specifically, the only extra
trainable parameters that our approach postulates stem from Eq. (17);
these are of extremely limited size compared to the overall model
size, and correspond to merely few extra feedforward computations
at inference time. Besides, our sampling strategy does not induce
significant computational costs, since we adopt the reparameterization
(\ref{approxsampling}) for the most part of the model training algorithm.
In the future, we aim to consider how ACVI can cope with power-law
distributions \cite{tnnls12,tnnls15}; such a capacity is of importance
to real-world natural language generation.

\captionsetup{justification=centering,singlelinecheck=false}

\begin{table*}[htp]
\caption{Example 223.}
\begin{tabularx}{0.99\textwidth}{|X|} \hline \textbf{Article}\tabularnewline
\hline \hline {lagos , nigeria -lrb- cnn -rrb- a day after winning
nigeria 's presidency , \textcolor{green}{muhammadu} \textcolor{green}{buhari
}told \textcolor{green}{cnn 's christiane amanpour} that he plans to
aggressively fight corruption that has long plagued nigeria and go
after the root of the \textcolor{green}{nation 's unrest} . buhari
said he 'll `` rapidly give attention '' to curbing violence in
the northeast part of nigeria , where the terrorist group boko haram
operates . by cooperating with neighboring nations chad , cameroon
and niger , he said his administration is confident it will be able
to thwart criminals and others contributing to nigeria 's instability
. \textcolor{green}{for the first time in nigeria} 's history , the
\textcolor{green}{opposition defeated} \textcolor{green}{the} ruling
party in democratic elections . buhari defeated \textcolor{green}{incumbent
goodluck jonathan by about 2 million votes} , according to nigeria
's independent national electoral commission . the win comes after
a long history of military rule , coups and botched attempts at democracy
in africa 's most populous nation . in an exclusive live interview
from abuja , buhari told amanpour he was not concerned about reconciling
the nation after a divisive campaign . he said now that he has been
elected he will turn his focus to boko haram and `` plug holes ''
in the `` corruption infrastructure '' in the country . `` a new
day and a new nigeria are upon us , '' \textcolor{green}{buhari} said
after his win tuesday . \textcolor{green}{`` the victory is yours
, and the glory is that of our nation . ''} earlier , jonathan phoned
buhari to concede defeat . the outgoing president also offered a written
statement to his nation . `` i thank all nigerians once again for
the great opportunity i was given to lead this country , and assure
you that i will continue to do my best at the helm of national affairs
until the end of my tenure , '' jonathan said . `` i promised the
country free and fair elections . (...)}\tabularnewline \hline
\textbf{Reference Summary}\tabularnewline \hline muhammadu buhari
tells cnn 's christiane amanpour that he will fight corruption in
nigeria . nigeria is the most populous country in africa and is grappling
with violent boko haram extremists . nigeria is also africa 's biggest
economy , but up to 70 \% of nigerians live on less than a dollar
a day . \tabularnewline \hline \textbf{Generated Summary}\tabularnewline
\hline \textit{muhammadu buhari} \textcolor{purple}{talks} to cnn
's christiane amanpour \textbf{about the nation 's unrest} . for the
first time in nigeria , opposition defeated incumbent goodluck jonathan
by about 2 million votes. \textit{buhari} : '' the victory is yours
, and the glory is that of our nation '' \tabularnewline \hline
\end{tabularx}
\end{table*}

\begin{table*}[htp]
\caption{Example 89.}
\begin{tabularx}{0.99\textwidth}{|X|} \hline \textbf{Article}\tabularnewline
\hline \hline {lrb- cnn -rrb- eyewitness \textcolor{green}{video}
showing \textcolor{green}{white north charleston police officer michael
slager shooting to death }an unarmed black man has exposed discrepancies
in the reports of the first officers on the scene . \textcolor{green}{slager
has been} fired and \textcolor{green}{charged with murder in the death
of 50-year-old walter scott} . a bystander 's cell phone video , which
began after an alleged struggle on the ground between slager and scott
, shows the five-year police veteran shooting at scott eight times
as scott runs away . scott was hit five times . if words were exchanged
between the men , they 're are not audible on the tape . it 's unclear
what happened before scott ran , or why he ran . the officer initially
said that he used a taser on scott , who , slager said , tried to
take the weapon . before slager opens fire , \textcolor{green}{the
video shows a dark object falling behind scott and hitting the ground}
. it 's unclear whether that is the taser . (...)}\tabularnewline
\hline \textbf{Reference Summary}\tabularnewline \hline more questions
than answers emerge in controversial s. c. police shooting . officer
michael slager , charged with murder , was fired from the north charleston
police department . \tabularnewline \hline \textbf{Generated Summary}\tabularnewline
\hline video \textbf{\textcolor{black}{shows}} white north charleston
police officer michael \emph{slager} shooting to death . \emph{slager}
has been charged with murder in the death of 50-year-old walter scott
. the video shows a dark object falling behind scott and hitting the
ground . \tabularnewline \hline \end{tabularx}
\end{table*}

\begin{table*}[htp]
\caption{Example 1305.}
\begin{tabularx}{0.99\textwidth}{|X|} \hline \textbf{Article}\tabularnewline
\hline \hline {andy murray came close to giving himself some extra
preparation time for his wedding next week before ensuring that he
still has unfinished tennis business to attend to . \textcolor{green}{the
world no 4 is into the semi-finals of the miami open} , but not before
getting a scare from 21 year-old austrian dominic thiem , who pushed
him to 4-4 in the second set before going down 3-6 6-4 , 6-1 in an
hour and three quarters . \textcolor{green}{murray was awaiting the
winner from the last eight match} between tomas berdych and argentina
's juan monaco . prior to this tournament thiem lost in the second
round of a challenger event to soon-to-be new brit aljaz bedene .
andy murray pumps his first after defeating dominic thiem to reach
the miami open semi finals . \textcolor{green}{muray throws his sweatband
into the crowd after completing a 3-6 , 6-4 , 6-1 victory in florida
. }murray shakes hands with thiem who he described as a ` strong guy
' after the game . (...)}\tabularnewline \hline \textbf{Reference
Summary}\tabularnewline \hline british no 1 defeated dominic thiem
in miami open quarter finals . andy murray celebrated his 500th career
win in the previous round . third seed will play the winner of tomas
berdych and juan monaco in the semi finals of the atp masters 1000
event in key biscayne \tabularnewline \hline \textbf{Generated Summary}\tabularnewline
\hline the world no 4 is into the semi-finals of the miami open .
\textbf{\emph{murray}}\textbf{ is still }\textbf{\textcolor{purple}{ahead}}\textbf{
of his }\textbf{\textcolor{purple}{career}}\textbf{ through the season
}. \textbf{andy} \emph{murray} was awaiting the winner from the last
eight match . \emph{murray} throws his sweatband into the crowd after
a 6-4 6-1 victory in florida . \tabularnewline \hline \end{tabularx}
\end{table*}

\begin{table*}[htp]
\caption{Example 1710.}
\begin{tabularx}{0.99\textwidth}{|X|} \hline \textbf{Article}\tabularnewline
\hline \hline {steve clarke afforded himself a few smiles on the
touchline and who could blame him ? this has been a strange old season
for reading , who are one win away from an fa cup semi-final against
arsenal but have spent too long being too close to a championship
relegation battle . at least this win will go some way to easing that
load . they made it hard for themselves , but they had an in-form
player in jamie mackie who was able to get the job done . he put reading
in front in the first half and then scored a brilliant winner just
moments after chris o'grady had levelled with a penalty -{}- one
of the only legitimate chances brighton had all night , even if clarke
was angry about the decision . reading frontman \textcolor{green}{jamie
mackie} fires the royals \textcolor{green}{ahead against brighton in
tuesday 's championship fixture} . mackie -lrb- centre -rrb- is congratulated
by nathaniel chalobah and \textcolor{green}{garath mccleary }after
\textbf{\textcolor{green}{netting}} reading 's opener . reading -lrb-
4-1-3-2 -rrb- : federici ; gunter , hector , cooper , chalobah ; akpan
; mcleary , williams -lrb- keown 92 -rrb- , robson-kanu -lrb- pogrebnyak
76 -rrb- ; blackman , mackie -lrb- norwood 79 -rrb- . subs not used
: cox , yakubu , andersen , taylor . scorer : mackie , 24 , 56 . booked
: mcleary , pogrebnyak . brighton -lrb- 4-3-3 -rrb- : stockdale ;
halford , greer , dunk , bennett ; ince -lrb- best 75 -rrb- , kayal
, forster-caskey ; ledesma -lrb- bruno 86 -rrb- , o'grady , lualua
. subs not used : ankergren , calderon , hughes , holla , teixeira
. \textbf{\textcolor{green}{scorer :}}\textcolor{green}{{} o'grady}
-lrb- pen -rrb- , 53 . booked : ince , dunk , bennett , greer . ref
: andy haines . attendance : 14,748 . ratings by riath al-samarrai
. (...) }\tabularnewline \hline \textbf{Reference Summary}\tabularnewline
\hline reading are now 13 points above the championship drop zone
. frontman jamie mackie scored twice to earn royals all three points
. chris o'grady scored for chris hughton 's brighton from the penalty
spot . niall keown - son of sportsmail columnist martin - made reading
debut . \tabularnewline \hline \textbf{Generated Summary}\tabularnewline
\hline \textbf{jamie }\textbf{\emph{mackie}}\textbf{ }\textbf{\textcolor{purple}{opened}}\textbf{
the scoring against brighton in tuesday 's championship fixture }.
chris \emph{o'grady} and \emph{garath} \emph{mccleary} \textcolor{purple}{both}
scored . \textbf{jamie }\textbf{\emph{mackie}}\textbf{ and}\textbf{\emph{
garath mccleary}}\textbf{ }\textbf{\textcolor{purple}{were both involved}}\textbf{
in the game }. \tabularnewline \hline \end{tabularx}
\end{table*}

\captionsetup{justification=centering,singlelinecheck=false}

\begin{figure*}[htp]
\begin{minipage}[c]{0.25\textwidth}%
\centering \includegraphics[width=3cm]{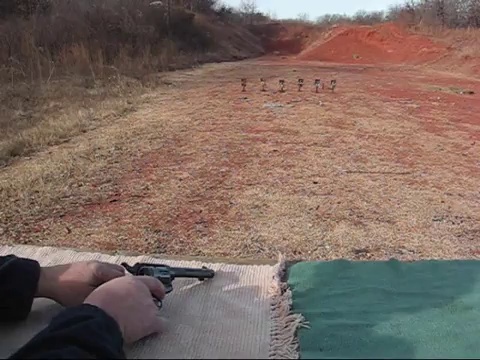} %
\end{minipage}%
\begin{minipage}[c]{0.25\textwidth}%
\centering \includegraphics[width=3cm]{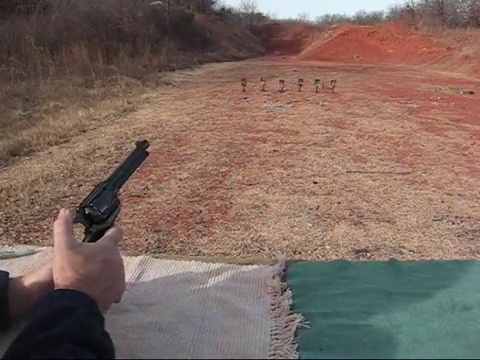} %
\end{minipage}%
\begin{minipage}[c]{0.25\textwidth}%
\centering \includegraphics[width=3cm]{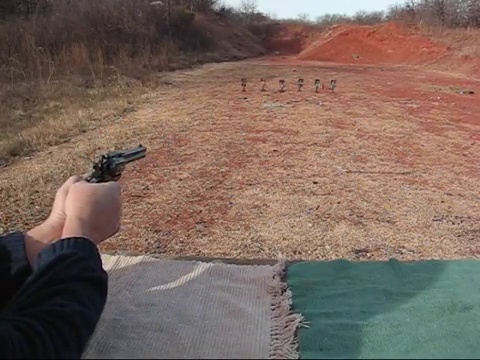} %
\end{minipage}%
\begin{minipage}[c]{0.25\textwidth}%
\centering \includegraphics[width=3cm]{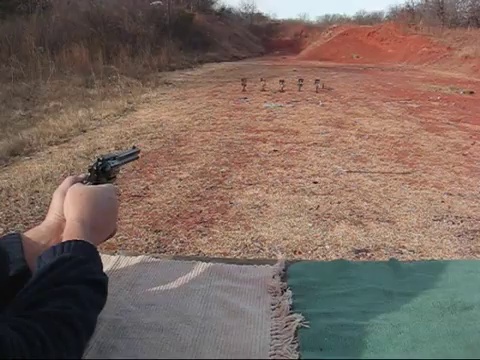} %
\end{minipage}\caption{\textbf{ACVI:} a man is firing a gun\protect \protect \protect \\
\textbf{VED:} a man is firing a gun\protect \protect
\protect \\
\textbf{Structured Attention:} a man is firing a gun\protect \protect
\protect \\
\textbf{Variational Attention:} a man is firing a gun\protect \protect
\protect \\
 \textbf{SA:} a man is firing a gun\protect \protect \protect \\
 \textbf{\-\hspace{3.1cm}Reference Description:} a man is firing
a gun \textcolor{blue}{at targets}}
\label{fig:1} 
\end{figure*}

\begin{figure*}[htp]
\begin{minipage}[c]{0.25\textwidth}%
\centering \includegraphics[width=3cm]{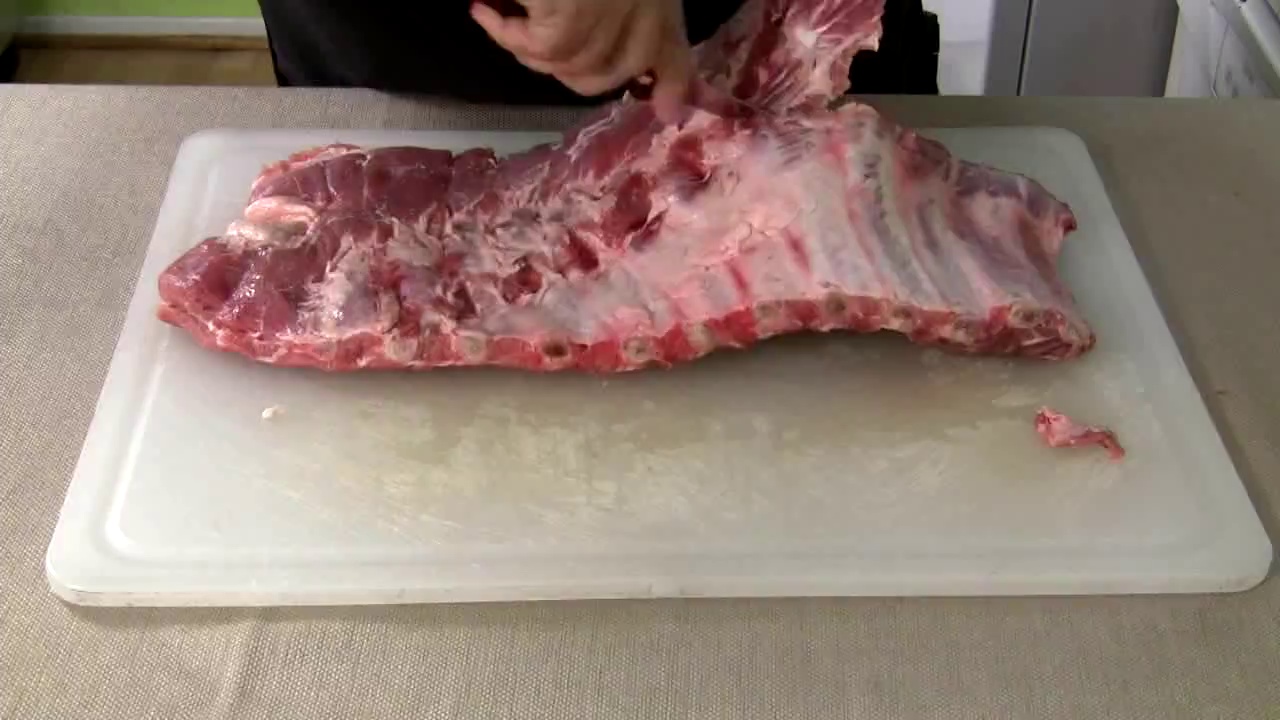} %
\end{minipage}%
\begin{minipage}[c]{0.25\textwidth}%
\centering \includegraphics[width=3cm]{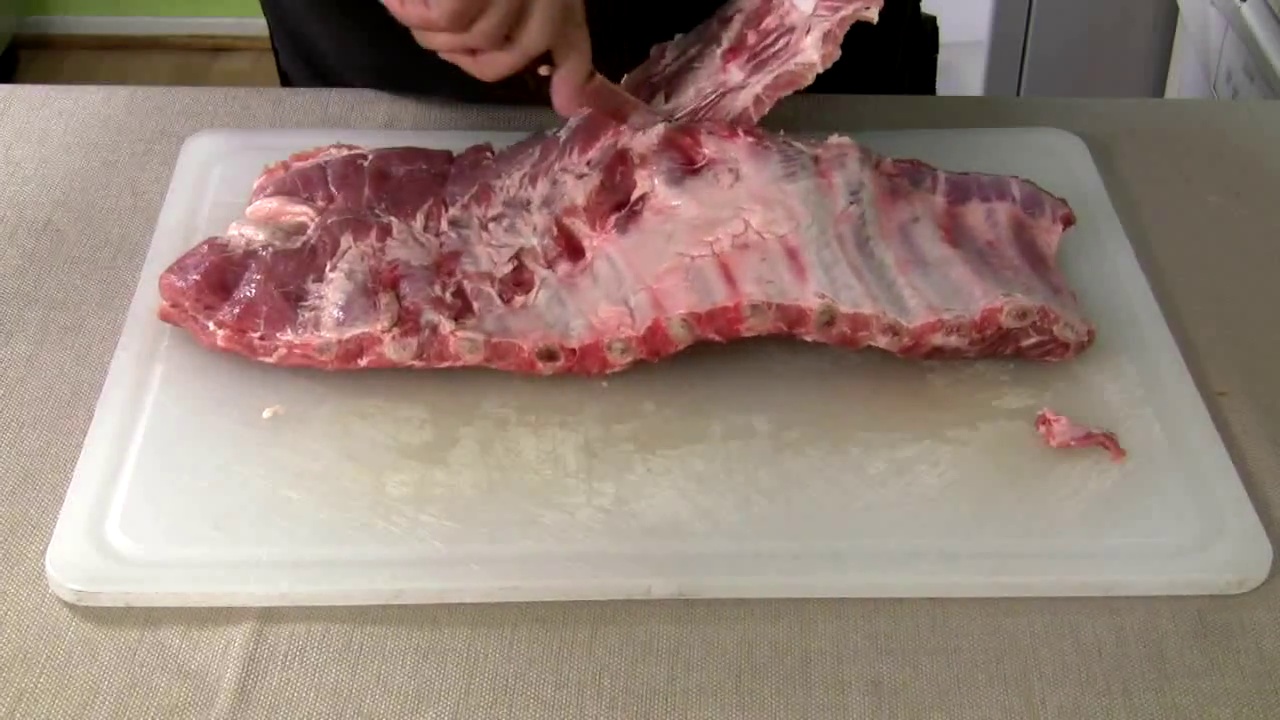} %
\end{minipage}%
\begin{minipage}[c]{0.25\textwidth}%
\centering \includegraphics[width=3cm]{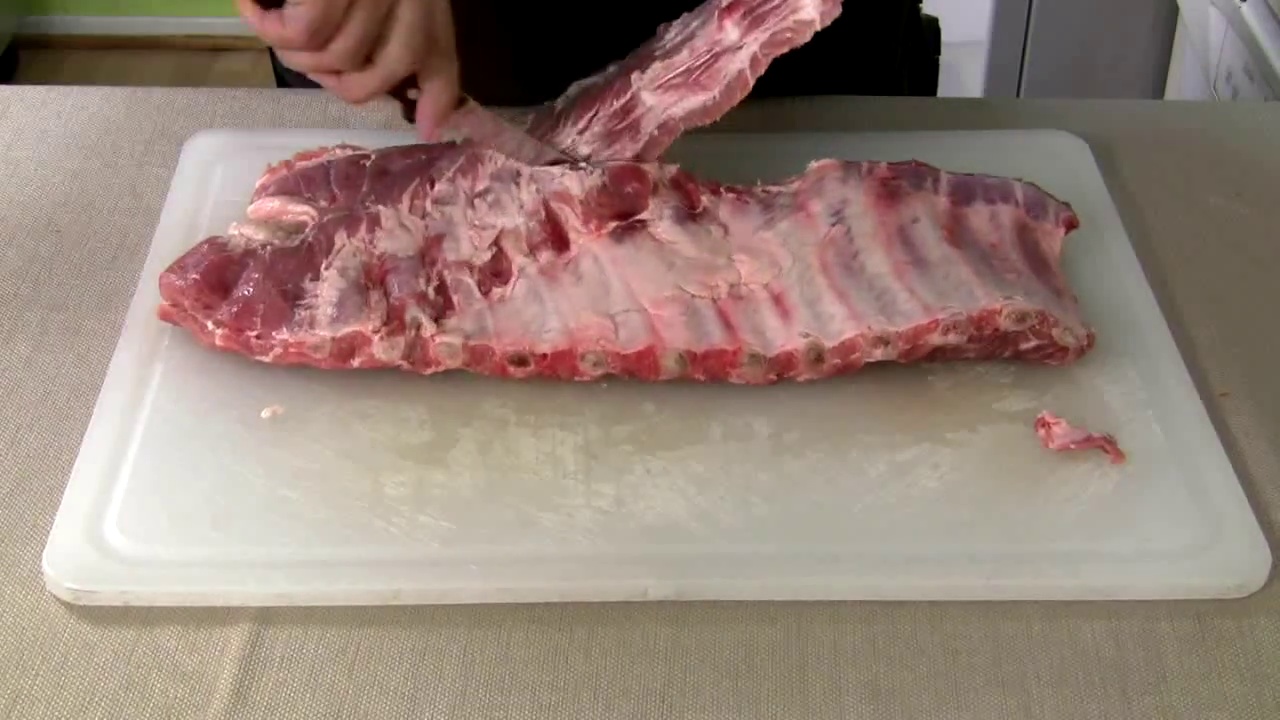} %
\end{minipage}%
\begin{minipage}[c]{0.25\textwidth}%
\centering \includegraphics[width=3cm]{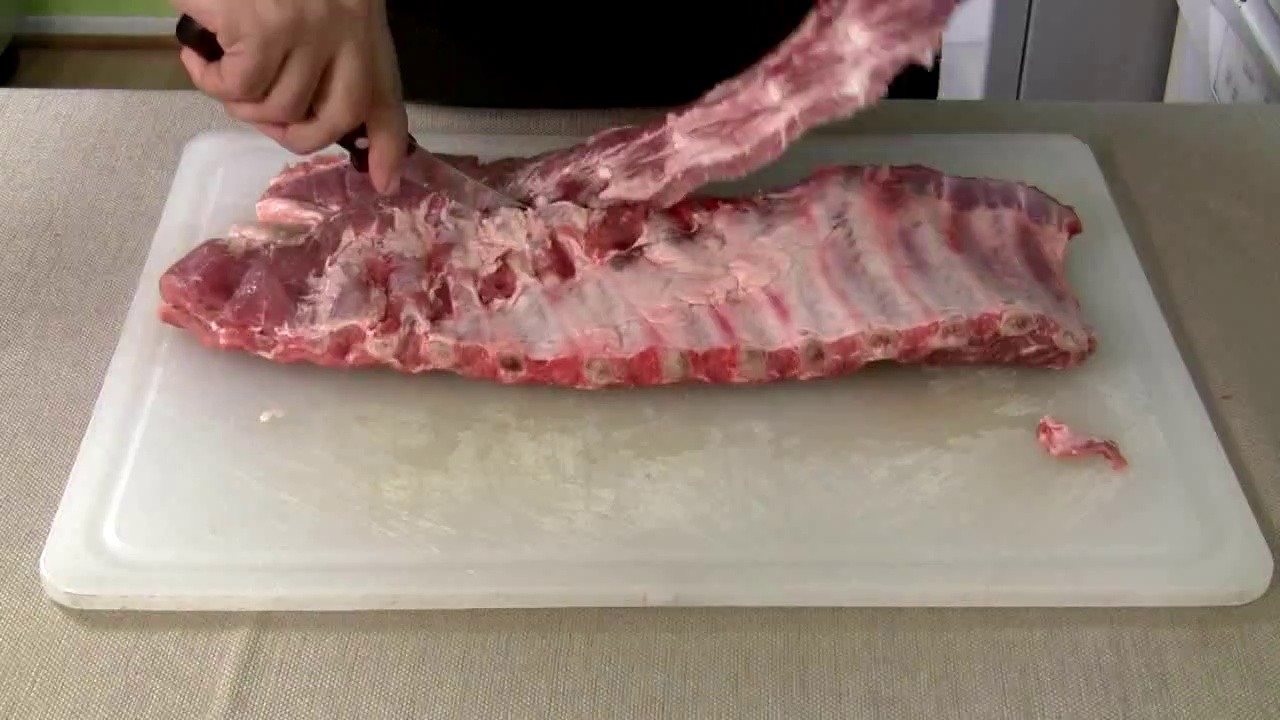} %
\end{minipage}\caption{\textbf{ACVI:} \textcolor{blue}{a woman} is cutting a piece of \textcolor{blue}{pork}\protect
\protect \protect \\
\textbf{VED:} \textcolor{blue}{a woman} is cutting \textcolor{red}{a bed}\protect \protect \\
\textbf{Structured Attention:} \textcolor{blue}{a woman} is cutting
\textcolor{blue}{pork}\protect \protect \protect \\
\textbf{Variational Attention:} \textcolor{blue}{a woman} is cutting
\textcolor{blue}{pork}\protect \protect \protect \\
 \textbf{SA:} \textcolor{blue}{a woman} \textcolor{red}{is putting
butter on a bed}\protect \protect \\
 \textbf{\-\hspace{1.8cm}Reference Description:} someone is cutting
a piece of meat}
\label{fig:2} 
\end{figure*}

\begin{figure*}[htp]
\begin{minipage}[c]{0.25\textwidth}%
\centering \includegraphics[width=3cm]{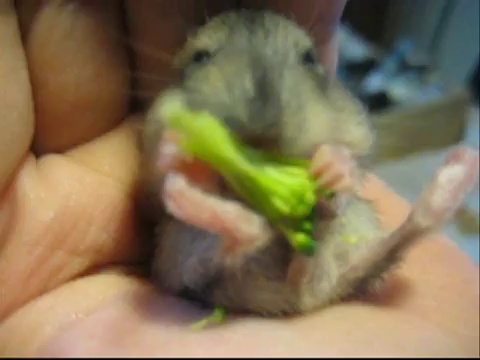} %
\end{minipage}%
\begin{minipage}[c]{0.25\textwidth}%
\centering \includegraphics[width=3cm]{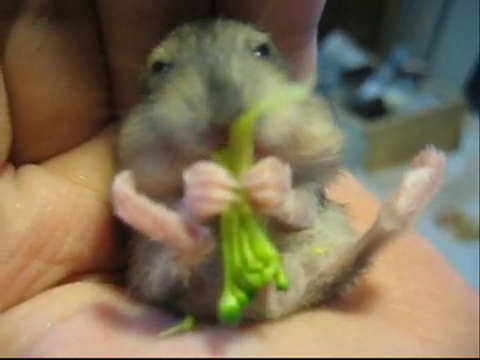} %
\end{minipage}%
\begin{minipage}[c]{0.25\textwidth}%
\centering \includegraphics[width=3cm]{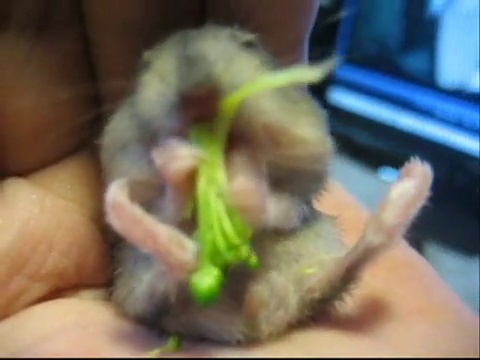} %
\end{minipage}%
\begin{minipage}[c]{0.25\textwidth}%
\centering \includegraphics[width=3cm]{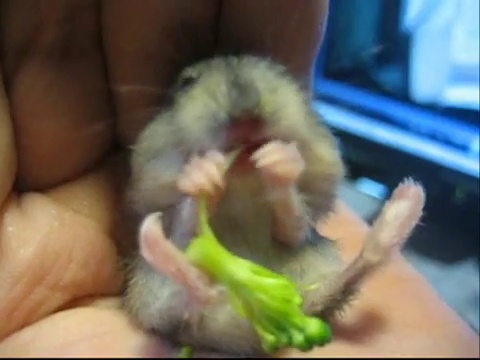} %
\end{minipage}\caption{\textbf{ACVI:} a \textcolor{blue}{small animal} is eating\protect
\protect \protect \\
 \textbf{VED:} a \textcolor{red}{small woman} is \textcolor{red}{talking}
\protect \protect \\
\textbf{Structured Attention:} a \textcolor{blue}{small }\textcolor{red}{woman}
is eating\protect \protect\protect \\
\textbf{Variational Attention:} a \textcolor{blue}{small }\textcolor{red}{woman}
is eating\protect \protect\protect \\
 \textbf{SA:} a \textcolor{red}{small woman} is \textcolor{red}{talking}
\protect \protect \\
 \textbf{\-\hspace{1.1cm}Reference Description:} a hamster is eating}
\label{fig:3} 
\end{figure*}

\begin{figure*}[htp]
\begin{minipage}[c]{0.25\textwidth}%
\centering \includegraphics[width=3cm]{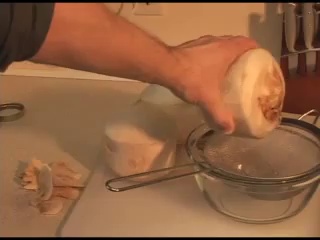} %
\end{minipage}%
\begin{minipage}[c]{0.25\textwidth}%
\centering \includegraphics[width=3cm]{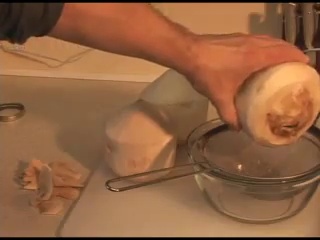} %
\end{minipage}%
\begin{minipage}[c]{0.25\textwidth}%
\centering \includegraphics[width=3cm]{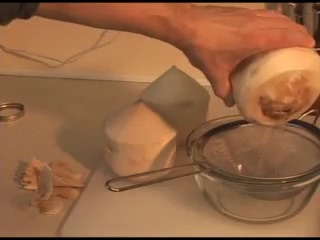} %
\end{minipage}%
\begin{minipage}[c]{0.25\textwidth}%
\centering \includegraphics[width=3cm]{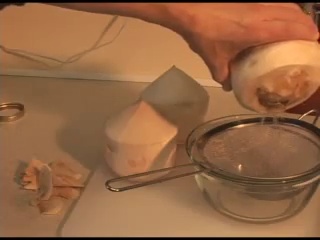} %
\end{minipage}\caption{\textbf{ACVI:} \textcolor{blue}{the lady poured} \textcolor{red}{the}
something into a bowl\protect \protect \protect \\
 \textbf{VED:} \textcolor{blue}{a woman} \textcolor{red}{is cracking
an egg}\protect \protect \\
\textbf{Structured Attention:} \textcolor{blue}{a woman poured} \textcolor{red}{an
egg} into a bowl\protect \protect \protect \\
\textbf{Variational Attention:} \textcolor{blue}{a woman poured} \textcolor{red}{an
egg} into a bowl\protect \protect \protect \\
 \textbf{SA:} \textcolor{blue}{a woman} \textcolor{red}{is cracking
an egg}\protect \protect \\
 \textbf{\-\hspace{1.8cm}Reference Description:} someone is pouring
something into a bowl}
\label{fig:4} 
\end{figure*}

\begin{figure*}[htp]
\begin{minipage}[c]{0.25\textwidth}%
\centering \includegraphics[width=3cm]{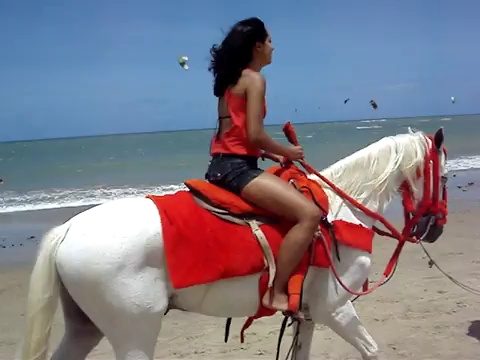} %
\end{minipage}%
\begin{minipage}[c]{0.25\textwidth}%
\centering \includegraphics[width=3cm]{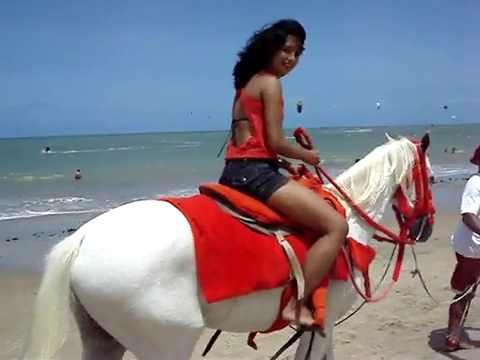} %
\end{minipage}%
\begin{minipage}[c]{0.25\textwidth}%
\centering \includegraphics[width=3cm]{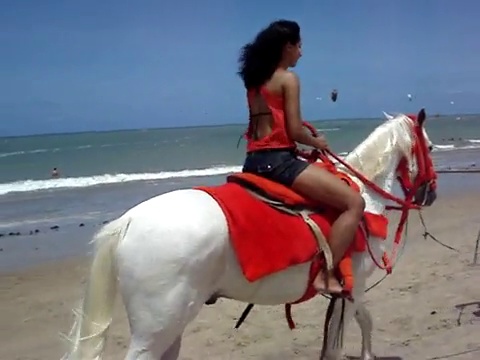} %
\end{minipage}%
\begin{minipage}[c]{0.25\textwidth}%
\centering \includegraphics[width=3cm]{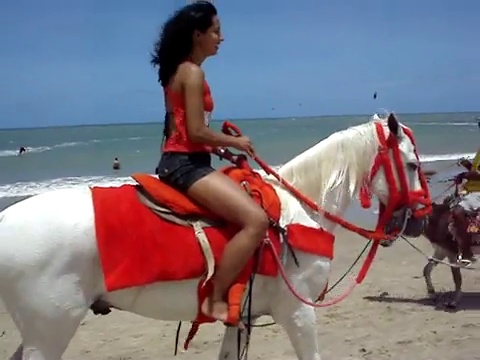} %
\end{minipage}\caption{\textbf{ACVI:} a woman is riding a horse\protect \protect \protect \\
\textbf{VED:} a woman is riding a horse\protect
\protect \protect \\
\textbf{Structured Attention:} a woman is riding a horse\protect
\protect \protect \\
\textbf{Variational Attention:} a woman is riding a horse\protect
\protect \protect \\
 \textbf{SA:} a woman is riding a horse \protect \protect \\
 \textbf{\-\hspace{1.7cm}Reference Description:} a woman is riding
a horse}
\label{fig:5} 
\end{figure*}

\begin{figure*}[htp]
\begin{minipage}[c]{0.25\textwidth}%
\centering \includegraphics[width=3cm]{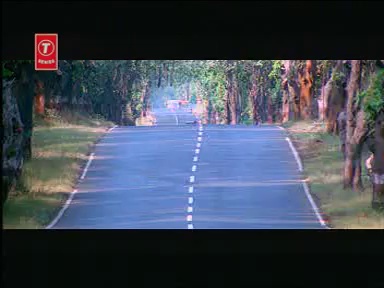} %
\end{minipage}%
\begin{minipage}[c]{0.25\textwidth}%
\centering \includegraphics[width=3cm]{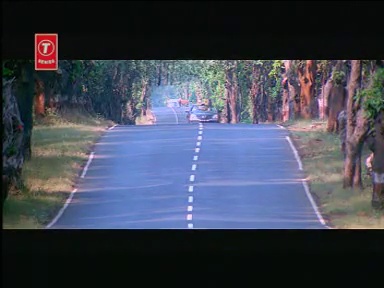} %
\end{minipage}%
\begin{minipage}[c]{0.25\textwidth}%
\centering \includegraphics[width=3cm]{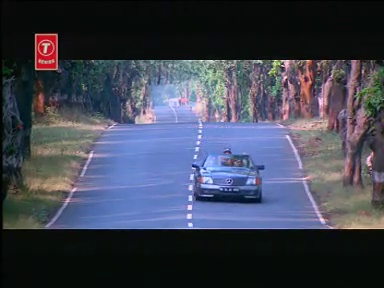} %
\end{minipage}%
\begin{minipage}[c]{0.25\textwidth}%
\centering \includegraphics[width=3cm]{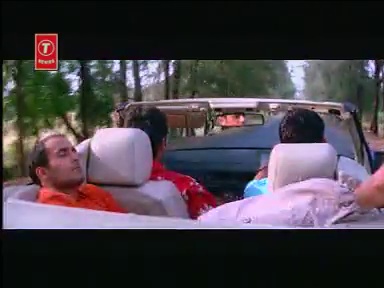} %
\end{minipage}\caption{\textbf{ACVI:} \textcolor{blue}{several people} are driving down a
\textcolor{blue}{street}\protect \protect \protect \\
\textbf{VED:} \textcolor{blue}{several people} \textcolor{red}{trying to jump}
\protect \protect \protect \\
\textbf{Structured Attention:} \textcolor{blue}{several people} are
driving down the \textcolor{red}{avenue}\protect \protect \protect \\
\textbf{Variational Attention:} \textcolor{blue}{several people} are
driving down the \textcolor{red}{avenue}\protect \protect \protect \\
 \textbf{SA:} \textcolor{red}{a boy trying to jump}\protect \protect \\
 \textbf{\-\hspace{0.4cm}Reference Description:} a car is driving
down the road}
\label{fig:6} 
\end{figure*}

\begin{figure*}[htp]
\begin{minipage}[c]{0.25\textwidth}%
\centering \includegraphics[width=3cm]{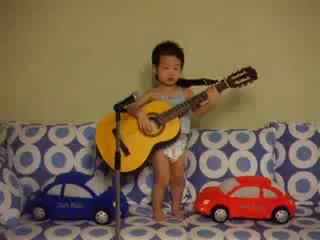} %
\end{minipage}%
\begin{minipage}[c]{0.25\textwidth}%
\centering \includegraphics[width=3cm]{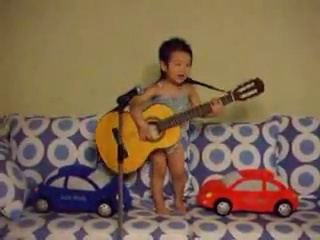} %
\end{minipage}%
\begin{minipage}[c]{0.25\textwidth}%
\centering \includegraphics[width=3cm]{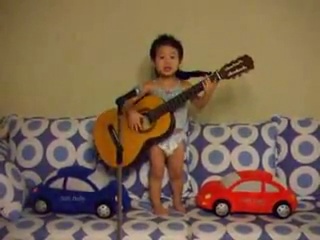} %
\end{minipage}%
\begin{minipage}[c]{0.25\textwidth}%
\centering \includegraphics[width=3cm]{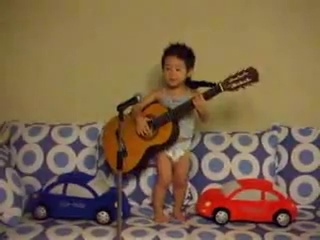} %
\end{minipage}\caption{\textbf{ACVI:} a \textcolor{blue}{man} is playing the guitar\protect
\protect \protect \\
\textbf{VED:} a \textcolor{blue}{{} man} is
\textcolor{red}{dancing} \protect \protect \\
\textbf{Structured Attention:} a \textcolor{red}{high}\textcolor{blue}{{}
man} is playing the guitar\protect \protect \protect \\
 \textbf{Variational Attention:} a \textcolor{blue}{{} man} is
\textcolor{red}{dancing} \protect \protect \\
 \textbf{SA:} a \textcolor{red}{high}\textcolor{blue}{{} man} is
\textcolor{red}{dancing} \protect \protect \\
 \textbf{\-\hspace{1.7cm}Reference Description:} a boy is playing
the guitar}
\label{fig:7} 
\end{figure*}

\begin{figure*}[htp]
\begin{minipage}[c]{0.25\textwidth}%
\centering \includegraphics[width=3cm]{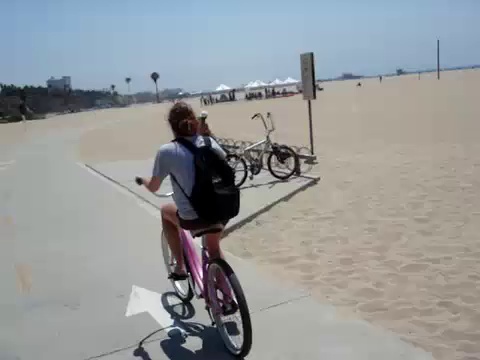} %
\end{minipage}%
\begin{minipage}[c]{0.25\textwidth}%
\centering \includegraphics[width=3cm]{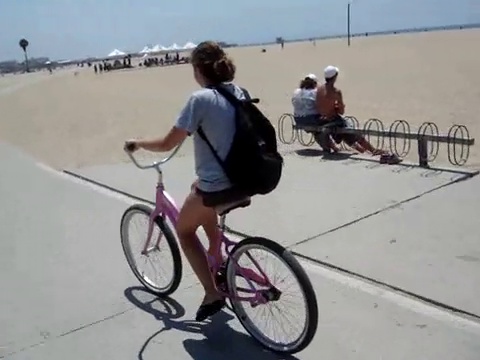} %
\end{minipage}%
\begin{minipage}[c]{0.25\textwidth}%
\centering \includegraphics[width=3cm]{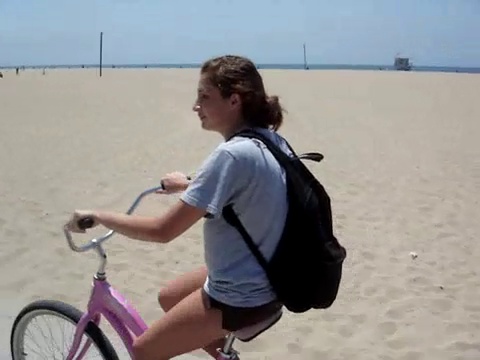} %
\end{minipage}%
\begin{minipage}[c]{0.25\textwidth}%
\centering \includegraphics[width=3cm]{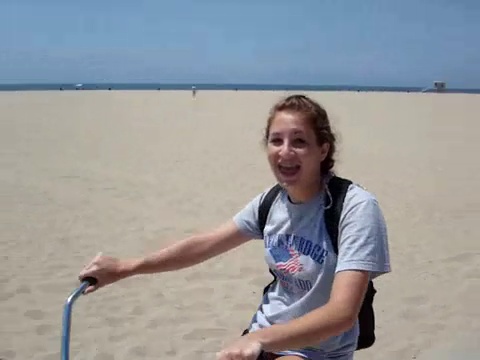} %
\end{minipage}\caption{\textbf{ACVI:} \textcolor{blue}{the man} is riding a bicycle\protect
\protect \protect \\
\textbf{VED:} \textcolor{blue}{the man} is riding
a \textcolor{red}{motorcycle}\protect \protect \protect \\
\textbf{Structured Attention:} \textcolor{blue}{the man} is riding
a \textcolor{red}{motorcycle}\protect \protect \protect \\
\textbf{Variational Attention:} \textcolor{blue}{the man} is riding
a \textcolor{red}{motorcycle}\protect \protect \protect \\
 \textbf{SA:} \textcolor{blue}{a man} \textcolor{red}{rides} a \textcolor{red}{motorcycle}
\protect \protect \\
 \textbf{\-\hspace{1.4cm}Reference Description:} a girl is riding
a bicycle}
\label{fig:8} 
\end{figure*}

\section*{Appendix A}

We elaborate here on the experimental setup of Section IV.A.  The used dataset comprises 287,226 training pairs of documents and reference summaries, 13,368 validation pairs,
and 11,490 test pairs. In this dataset, the average article length
is 781 tokens; the average summary length is 3.75 sentences, with
the average summary being 56 tokens long. In all our experiments,
we restrict the used vocabulary to the 50K most common words in the
considered dataset, similar to \cite{Manning2017}. Note that this is significantly smaller than typical in the literature \cite{Nallapati2016}.

To allow for faster training convergence, we split it into five phases.
On each phase, we employ a different number of maximum encoding steps
for the evaluated models (i.e., the size of the inferred attention
vectors), as well as for the maximum allowed number of decoding steps.
We provide the related details in Table 5. During these phases, we
train the employed models with the coverage mechanism being disabled;
that is, we set $\boldsymbol{w}_{k}=\boldsymbol{0}$. We enable this
mechanism only after these five training phases conclude. Specifically,
we perform a final 3K iterations of model training, during which we
train the $\boldsymbol{w}_{k}$ weights along with the rest of the
model parameters. We do \emph{not} use any form of regularization, as suggested
in \cite{Manning2017}.

In Tables 6-9, we provide some indicative examples of summaries produced by
a pointer-generator network with coverage, employing the ACVI mechanism.
We also show what the initial document has been, as well as the available
reference summary used for quantitative performance evaluation. In
all cases, we annotate OOV words in italics, we highlight novel words
in purple, we show contextual understanding in bold, while article
fragments also included in the generated summary are highlighted in
green.


\section*{Appendix B}

The considered Video Captioning task utilizes a dataset that comprises 1,970
video clips, each associated with multiple natural language descriptions.
This results in a total of approximately 80,000 video / description
pairs; the used vocabulary comprises approximately 16,000 unique words.
The constituent topics cover a wide range of domains, including sports,
animals and music. We split the available dataset into a training
set comprising the first 1,200 video clips, a validation set composed
of 100 clips, and a test set comprising the last 600 clips in the
dataset. We preprocess the available descriptions only using the \emph{wordpunct} \emph{tokenizer} from the NLTK toolbox ({\url{http:/s/www.nltk.org/index.html}). We perform Dropout regularization
of the employed LSTMs, as suggested in \cite{rnndropout}; we use
a dropout rate of 0.5.

We provide some characteristic examples of generated
video descriptions in Figs. 1-8. In the captions of the figures that follow, we
annotate minor deviations with blue color, and use
red color to indicate major mistakes which imply wrong perception
of the scene.

\section*{References}
\bibliographystyle{elsarticle-num}
\bibliography{avi}

\end{document}